\documentclass{svproc}
\usepackage{url}

\usepackage{microtype}
\usepackage{graphicx}
\usepackage{booktabs} 
\usepackage{subfigure}

\begin{document}
\mainmatter              
\title{Smooth Grad-CAM++: An Enhanced Inference Level Visualization Technique for Deep Convolutional Neural Network Models}
\titlerunning{Smooth Grad-CAM++ for  Deep CNN Visualization}  
%
\author{Daniel Omeiza, Skyler Speakman, Celia Cintas, Komminist Weldemariam}
%

%
\tocauthor{Daniel Omeiza, Skyler Speakman, Celia Cintas, Komminist Weldermariam}

\maketitle              

\begin{abstract}

Gaining insight into how deep convolutional neural network models perform image classification and how to explain their outputs have been a concern to computer vision researchers and decision makers. These deep models are often referred to as black box due to low comprehension of their internal workings. As an effort to developing explainable deep learning models, several methods have been proposed such as finding gradients of class output with respect to input image (sensitivity maps), class activation map (CAM), and Gradient based Class Activation Maps (Grad-CAM). These methods under perform when localizing multiple occurrences of the same class and do not work for all CNNs. In addition, Grad-CAM does not capture the entire object in completeness when used on single object images, this affect performance on recognition tasks. With the intention to create an enhanced visual explanation in terms of visual sharpness, object localization and explaining multiple occurrences of objects in a single image, we present Smooth Grad-CAM++ \footnote{Simple demo: http://35.238.22.135:5000/}, a technique that combines methods from two other recent techniques---SMOOTHGRAD and Grad-CAM++. Our Smooth Grad-CAM++ technique provides the capability of either visualizing a layer, subset of feature maps, or subset of neurons within a feature map at each instance at the inference level (model prediction process). After  experimenting with few images, Smooth Grad-CAM++ produced more visually sharp maps with better localization of objects in the given input images when compared with other methods.
\keywords{Computer Vision, Convolutional Neural Network, Class Activation Maps}
\end{abstract}
\section{Introduction}
\label{submission}

Today, 
many deep learning models perform  well with good results in tasks like object detection, speech recognition, machine translation and few others \cite{b1}. According to \cite{b2},  breakthroughs are evident in some computer vision tasks which include image classification \cite{b3}, object detection \cite{b6}, semantic segmentation \cite{b7}, image captioning \cite{b4}, and visual question answering \cite{b5}. As much as the performance of these complex models are improved, there are limitations in the effectiveness of the conventional intelligent models as they lack the ability to explain their decisions to human users. This is a non-trivial issue in risk-averse domains such as in security, health and autonomous navigation \cite{b1}. 

More interpretable models can reveal important but surprising patterns in the data that complex models overlooked \cite{b8}. This was made evident in the pneumonia risk prediction case study in \cite{b9}. Artificial Intelligent agents are weaker than humans and not yet completely reliable. Thus, transparency and explainability is key in neural network models to identify failure modes \cite{b10}.
Narrowing down to image classification, few techniques have been proposed to understand the decisions of image classification models. A common approach usually called saliency (sensitivity or pixel attribution) is to find regions or subset of pixels of an image that were particularly influential to the final classification by the model \cite{b11,b12,b13}. 
This approach, including the general Class Activation Map (CAM) under perform when localizing multiple occurrences of the same class. Also, gradient based CAM (Grad-CAM) does not capture the entire object in completeness when used on single object images, this affect performance on recognition tasks. Although, Grad-CAM++ technique tends to take care of these limitations, improvements are required in terms of class object capturing, localization and visual appeal. In addition, the implementation of the current visualization techniques does not put into consideration the visualization of single or subset of neurons in a feature map, they only stop at the feature map level.

In this paper, we introduce gradient smoothening into Grad-CAM++, the resulting technique makes provision for visualizing a convolutional layer, subset of feature maps and subset of neurons in a feature map with improved visual appeal, localization and class object capturing. 
Smoothening entails adding noise to the sample image of interest, and  taking the average of all gradient matrices generated from each noised image. Grad-CAM++ does pixel-wise weighting of the gradients of the output with respect to a particular spatial position in the final convolutional feature map of the CNN. This provides a measure of importance of each pixel in a feature map towards the overall decision of the CNN.

\section{Background}
In this section, we discuss previous efforts put into understanding outputs from CNNs.
One of the earliest efforts in understanding deep CNNs is the deconvolution approach called Deconvnet. In this method, data flow from a neuron activation in the higher layers to lower layers, then parts of the image that highly influence that neuron are highlighted in the process \cite{b14}.
This led to the guided backpropagation idea. \cite{b15} introduced a new variant of the “deconvolution approach” called guided backpropagation for visualizing features learned by CNNs, which can be applied to a broader range of network structures than existing approaches. 
\cite{b16} presented a visualization toolbox to synthesize the input image that causes a specific unit in a neural network to have a high activation, this helps in visualizing the functionality of the unit. The toolbox could show activations for input images from a webcam or an image file, and gives intuition to what each filter is doing in each layer.
Class-specific saliency maps which are generated by performing a gradient ascent in pixel space to reach a maxima was proposed by \cite{b17}. This proves to be a more guided approach to synthesizing input images that maximally activate a neuron and helps to give better explanations on how a given CNN modeled a class \cite{b1}.
Other interesting approaches were proposed such as Local Interpretable Model-Agnostic Explanations(LIME) \cite{b18}, DeepLift \cite{b19}, and Contextual Explanation Networks (CENs) \cite{b20}.

More recent visualization techniques are Class Activation Map(CAM), Gradient-Weighted Class Activation Map(Grad-CAM) and a generalization of Grad-CAM called Grad-CAM++.  
In CAM, \cite{b11} demonstrated that a CNN with a Global Average Pooling (GAP) layer shows to have remarkable localization ability despite being trained on image-level labels. The CAM works for modified image classification CNNs that do not contain fully connected layers. The Grad-CAM is a generalization of the CAM for any CNN-based architecture.
While CAM is limited to a narrow class of CNN models, Grad-CAM is broadly applicable to any CNN-based architectures and needs no re-training.
The Grad-CAM technique
\begin{itemize}
    \item 
    computes the gradient of the class score(\(Y^{c}\)) with respect to feature map of the last convolution layer that is:
     \[ \frac{\partial Y^{c}}{\partial A^{K}_{i,j}}\]
     
    \item
    The gradients flowing back are global-averaged-pooled to obtain weights \(W^{c}_{k}\).
    \begin{equation}
    W^{c}_{k} = \frac{1}{Z} \sum_{i} \sum_{j} \frac{\partial Y^{c}}{\partial A^{K}_{i,j}}
    \label{eq:weight}
    \end{equation}
    Where \(W^{c}_{k}\) captures the importance of the feature map K for a target class c.
    
    \item 
    The Grad-CAM heatmap is a weighted combination of feature maps with ReLU.
    \begin{equation}
        L^{c}_{Grad-CAM} = ReLU\left(\sum_{k}{W^{c}_{k} A^{k}}\right)
        \label{eq:ReLU}
    \end{equation}
    
\end{itemize}

In Grad-CAM++, \cite{b1} proposed an algorithm which uses the weighted combination of the positive partial derivatives of the last convolutional layer feature maps with respect to a specific class score as weights to generate a visual explanation for the class label under consideration.
The author formulated:
\begin{itemize}
    \item 
    \(W^{c}_{k}\) to capture the importance of a particular activation map \(A_{k}\) by:
    \begin{equation}
        W^{c}_{k} = \sum_{i}\sum_{j}\alpha^{kc}_{i,j} ReLU\left(\frac{\partial Y^{c}}{\partial A^{k}_{i,j}}\right)
        \label{eq:activationmap}
    \end{equation}
    where \( \alpha^{kc}_{i,j} \) captures the importance of location (i,j) for activation map \(A^{k}\) for target class c.

    \item
    Replaced \(W^{c}_{k}\) in the fundamental assumption of CAM(not shown here) to give:
    \begin{equation}
        Y^{c} = \sum_{k}\left[\sum_{i}\sum_{j}\left(\sum_{a}\sum_{b} \alpha^{kc}_{a,b} ReLU\left(\frac{\partial Y^{c}}{\partial A^{k}_{i,j}}\right) A^{k}\right)\right]
        \label{eq:assumption}
    \end{equation}
    \item
Taking partial derivative with respect to \(A^{k}_{i,j}\) and rearranging terms gives the importance of each location in each activation map.\\
    \begin{equation}
        \alpha^{kc}_{i,j} = \frac{\frac{\partial^2 Y^{c}}{\left(\partial A^{k}_{i,j}\right)^2}}{2\frac{\partial^2 Y^{c}}{\left(\partial A^{k}_{i,j}\right)^2}+ \sum_{a}\sum_{b} A^{k}_{a,b} \frac{\partial^3 Y^{c}}{\left(\partial A^{k}_{i,j}\right)^3}}
        \label{eq:location}
    \end{equation}
    
\end{itemize}
\cite{b8} introduced SMOOTHGRAD, a simple method that can help visually sharpen gradient-based sensitivity maps by taking random samples in a neighborhood of an input x, and averaging the resulting sensitivity maps. Formally, this
means calculating:
\begin{equation}
    M_c\left(x\right) = \frac{1}{n}\sum_{1}^{n}M_{c}\left(x + \mathcal{N}\left(0, \sigma^2\right)\right)
    \label{eq:smoothgrad}
\end{equation}
where n is the number of samples, and \( \mathcal{N}\left(0, \sigma^2\right) \) represents Gaussian noise with standard deviation \(\sigma\).
With the motivation to provide an enhanced visualization maps, we apply this smoothening technique in the gradients computations involved in Grad-CAM++ as shown above. The resulting gradients are applied in the Grad-CAM++ algorithm. This provides better (in terms of visual appeal, localization and capturing) maps for deep CNNs.

\section{Method}
In this section we discuss the smoothening process (noise over input)and gradient averaging. We also discuss how the API was used to generate our results and how it could be used subsequently by other users. This involves model, convolution layer, feature map, and neuron selection. 
\subsection{Noise Over Input }
We set the number of noised sample images \(n\) to be generated by adding Gaussian noise to the original input. The standard deviation from mean value of the input is set, which provides the degree of noise to be added. We provide an API for interacting with the algorithm. The provided API uses 0 as the number of noised sample images to be generated by default. This implies that the gradients of the original input is used with no noise. The default standard deviation value is set to 0.15. These values could be varied till a satisfactory visual map is produced.
\subsection{Gradients Averaging}
We take the average of all 1st, 2nd and 3rd order partial derivatives of all \(n\) noised inputs and apply the resulting averaged derivatives in computing \(\alpha^{kc}_{i,j}\) and \( W^{c}_{k}\).\\

Let \(D_{1}^{k}, D_{2}^{k} \) and \(D_{3}^{k} \) denote matrices of 1st, 2nd and 3rd order partial derivatives respectively for feature map k.
We compute \(\alpha^{kc}\) as:
\begin{equation}
    \alpha^{kc}_{i,j} = \frac{\frac{1}{n}\sum_{1}^{n}D_{1}^{k}}{2\frac{1}{n}\sum_{1}^{n}D_{2}^{k} + \sum_{a}\sum_{b} A^{k}_{a,b} \frac{1}{n}\sum_{1}^{n}D_{3}^{k}}
    \label{eq:computed_a}
\end{equation}
substituting the averaged gradient into Equation \ref{eq:activationmap}, Grad-CAM++ weights \( W^{c}_{k}\) becomes:
\begin{equation}
    W^{c}_{k} = \sum_{i}\sum_{j}\alpha^{kc}_{i,j} ReLU\left(\frac{1}{n}\sum_{1}^{n}D_{1}^{k}\right)
    \label{eq:gradplus}
\end{equation}
When \(W^{c}_{k}\) is substituted into Equation \ref{eq:ReLU}, we get the final class discriminative saliency matrix which could be plotted with matplotlib or any other image plotting library. This serves as the final saliency map; this modified Grad-CAM++ is what we call Smooth Grad-CAM++.

\begin{figure*}[htbp]
\centerline{\includegraphics[width=12cm,height=7.5cm]{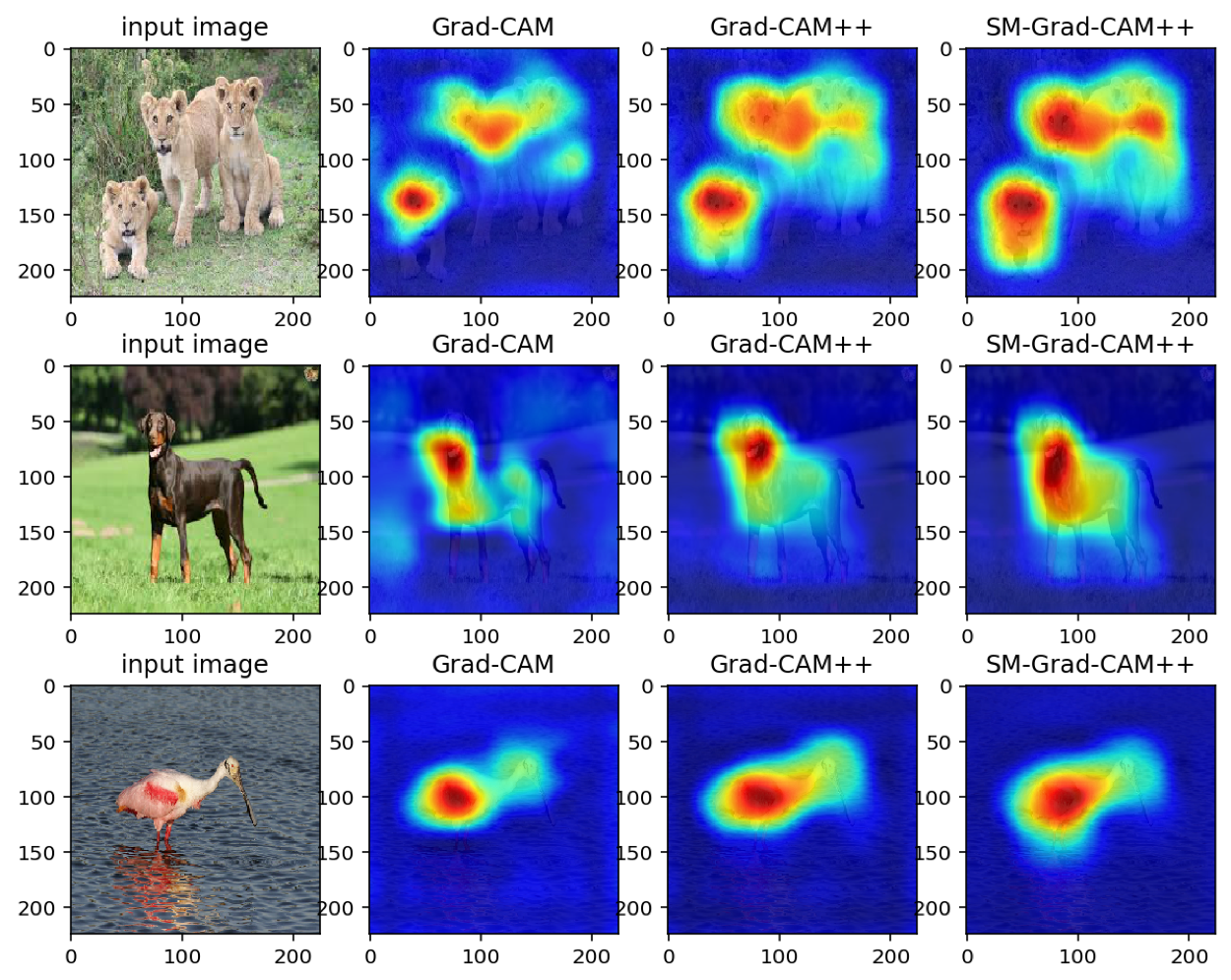}}
\caption{Saliency maps generated using Grad-CAM, Grad-CAM++ and Smooth Grad-CAM++ respectively, nsample=5, \(\sigma = 0.3\)}
\label{fig:layer1}
\end{figure*}



\begin{figure*}[htbp]
\centerline{\includegraphics[width=10cm, height=6.5cm]{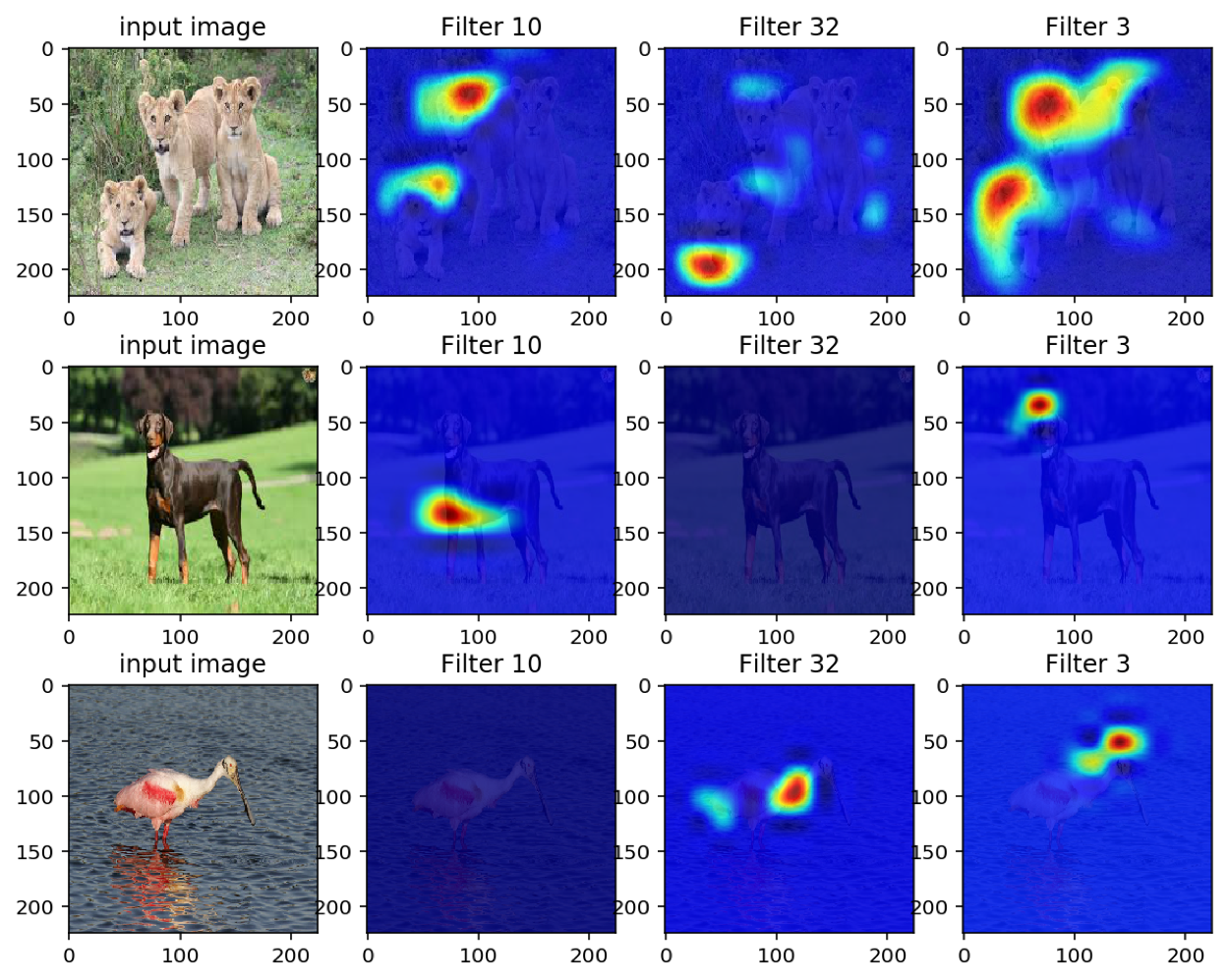}}
\caption{Saliency maps generated using Smooth Grad-CAM++ for specific feature map, nsample=5, \(\sigma = 0.3\)}
\label{fig:filter1}
\end{figure*}






\begin{figure*}[t]
\centerline{\includegraphics[width=7.5cm, height=9cm]{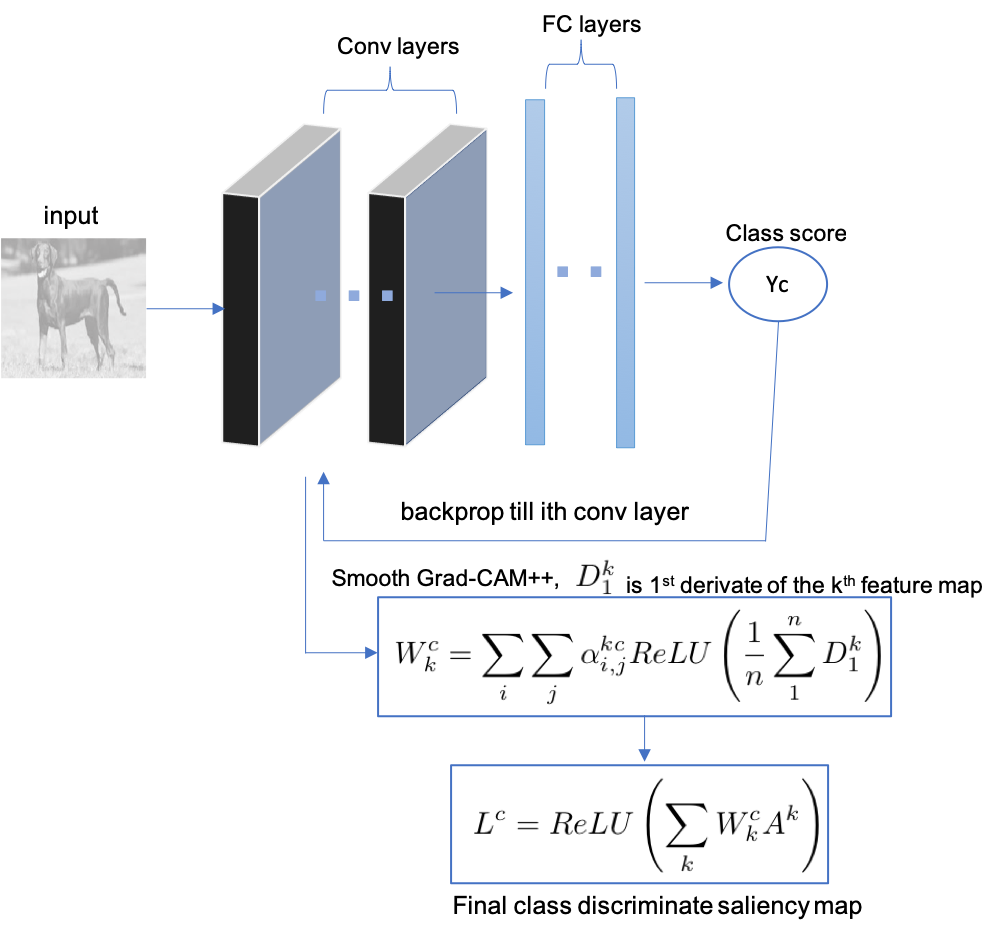}}
\caption{Overview of Smooth Grad-CAM++}
\label{fig}
\end{figure*}

\begin{figure}[t]
	\centering
	\includegraphics[width=7cm, height=4cm]{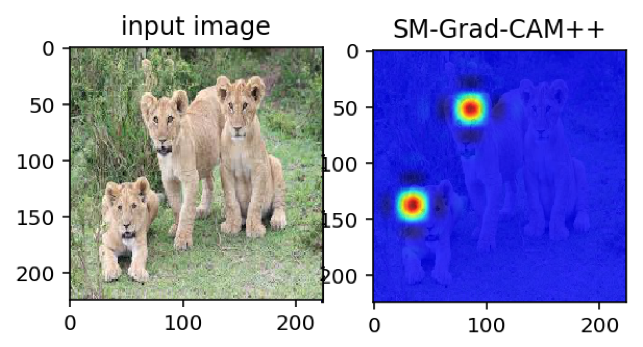}
	\caption{Neurons at coordinates (2,8) and (5,3) of feature map 3, nsample=5, \(\sigma = 0.3\)}
	\label{fig:node1}
\end{figure}

\begin{figure}[t]
	\centering
	\includegraphics[width=7cm, height=4cm]{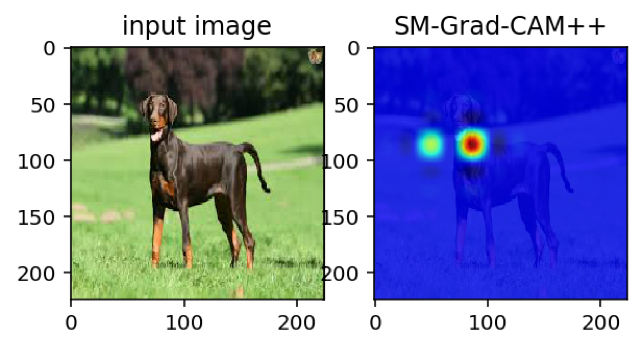}
	\caption{Neurons at coordinates (3,5) and (5,5) of feature map 10, nsample=5, \(\sigma = 0.3\)}
	\label{fig:node2}
\end{figure}

  



\subsection{Choosing a Model}
Any learned deep CNN model can be chosen for visualization. In this paper, we used VGG-16 pre-trained model and explored the last convolutional layer.

\subsection{Choosing Layer}
At each instance, only one convolution layer can be visualized. The name of the layer to be visualized is passed to the API. Names by default have a specific convention, however, viewing the summary of the trained model will reveal the name or unique identifier of each convolution layer. Each layer contains \(K\) feature maps which is usually with dimension of about
\[\left(\frac{H - F_{h} + 2P}{S_{h}} + 1, \frac{W - F_{w} + 2P}{S_{w}} + 1\right)\]
where \(H\) is the input height, \(W\) is the input width, \(P\) is the amount of zero padding added, \(F_{h}\) is the kernel height, \( F_{w}\) is the kernel width, \(S_{w}\) and \(S_{h}\) are horizontal and vertical strides of the convolutions respectively.
If no padding is supplied, then dimension would be about:
\[\left(ceil\left(\frac{H}{S_{h}}\right), ceil\left(\frac{W}{S_{w}}\right) \right)\]
where ceil rounds up the value to the nearest higher integer.
\subsection{Choosing Feature Maps}
To specify the feature maps to be visualized, the filter parameter must be set. The filter parameter is a list of integers specifying the index of the feature maps to be visualized in the specified convolution layer. If \(n\) values are set for the filter parameter, \(n\) maps are generated corresponding to each feature map.
For instance, if filter values are: \(0,1,2,3\), it means the values of \(k\) in the respective equations are bounded to \([0,3]\).

Recall that the number of feature maps corresponds to the number of kernels.
\subsection{Choosing Neurons}
One of the main contribution of this work is the capability of our technique to visualize subsets of neurons in a feature map. Visualizing neurons is useful when individual neuron activation is of interest. For instance, \cite{b21} used subset scan algorithm to identify anomalous activations in a convolutional neural network. Smooth Grad-CAM++ will be useful in providing explanations at the neuron level.
Our API provides an option to visualize regions of neurons within a specified coordinate boundary when region parameter is set to true. When region parameter is set to false and a subset of coordinates is provided, only the neurons in those coordinates are visualized while other activations are clipped at zero.
Smooth Grad-CAM++ could be a very flexible tool for debugging CNN models.

\subsection{API Call}
Necessary arguments are passed during API calls as shown below: 
\begin{verbatim}
grads = grad_cam_plus_smooth(model, img,
layer_name='block5_conv3', nsamples=5,
std_dev=0.3, filters=[0], region=True,
subset=[(0,10),(12, 12)])
\end{verbatim}
If the region parameter is set to true, the two coordinates in the subset would be treated as bound for neurons to be visualized. Hence, all neurons within the set bound are visualized. If region is false, each coordinate specified in the subset list is visualized.

\section{Results}
From Figure \ref{fig:layer1}, Smooth Grad-CAM++ gives a clearer explanation of particular features the model learned. For instance, Smooth Grad-CAM was able to highlight larger portion of the water-bird's legs in Figure \ref{fig:layer1}. Also, Smooth Grad-CAM++ captures larger amount of the class object (as seen in the dog image in Figure\ref{fig:layer1}), and does a good localization.
Figure \ref{fig:filter1} shows visual map for 3 randomly selected feature maps which are feature map 10, 32 and 3. Each feature map learns special features, some may be blank as seen in Figure \ref{fig:filter1}.
Figure \ref{fig:node1} and \ref{fig:node2} and show saliency map at neuron level for specific feature maps as captioned in the labels. This technique is a step towards gaining insights on what CNN models actually learn.

\section{Conclusion}
An enhanced visual saliency map can help increase our understanding of the internal workings of trained deep convolutional neural network models at the inference stage. In this paper, we proposed Smooth Grad-CAM++, an enhanced visual map for deep convolutional neural networks. Our results disclosed improvements in the generated visual maps when compared to existing methods. These maps were generated by averaging gradients (i.e derivative of class score with respect to the input) from many small perturbations of a given image and applying the resulting gradients in the generalized Grad-CAM algorithm (Grad-CAM++). Smooth Grad-CAM++ performs well in object localization and also in multiple occurrences of an object of same class. It is able to create maps for specific layers, subset of feature maps and neurons of interest. This will provide better insights on machine learning model explainability to machine learning researchers. Future works entail further investigations to extend this technique to handle multiple class scenarios, and different network architectures aside CNNs.
%

\end{document}